\documentclass{article}
\usepackage[nonatbib,preprint]{neurips}
\usepackage[utf8]{inputenc} %
\usepackage[T1]{fontenc}    %
\usepackage[pagebackref=true,breaklinks=true,colorlinks,bookmarks=false,citecolor=green,linkcolor=red]{hyperref}
\usepackage{url}            %
\usepackage{booktabs}       %
\usepackage{amsfonts}       %
\usepackage{nicefrac}       %
\usepackage{microtype}      %
\usepackage[table]{xcolor}  %
\usepackage{amsmath}
\usepackage{amssymb}
\usepackage{graphicx}
\usepackage{enumitem}
\usepackage[ruled,vlined]{algorithm2e}
\usepackage{wrapfig}
\usepackage{bm}
\usepackage{float}
\usepackage{subfig}
\usepackage{caption}
\usepackage{multirow}

\usepackage{array}
\newcolumntype{C}[1]{>{\centering\let\newline\\\arraybackslash\hspace{0pt}}m{#1}}
\usepackage{tablefootnote}

\title{\emph{SDXS}: Real-Time One-Step Latent Diffusion Models with Image Conditions}

\author{%
Yuda Song \quad\quad
Zehao Sun \quad\quad
Xuanwu Yin
\\
Xiaomi Inc.
\\
{\href{https://idkiro.github.io/sdxs/}{https://idkiro.github.io/sdxs/}}
}

\begin{document}

\maketitle

\begin{figure*}[h]
    \centering
    \vspace{-1.7em}
    \includegraphics[width=\linewidth]{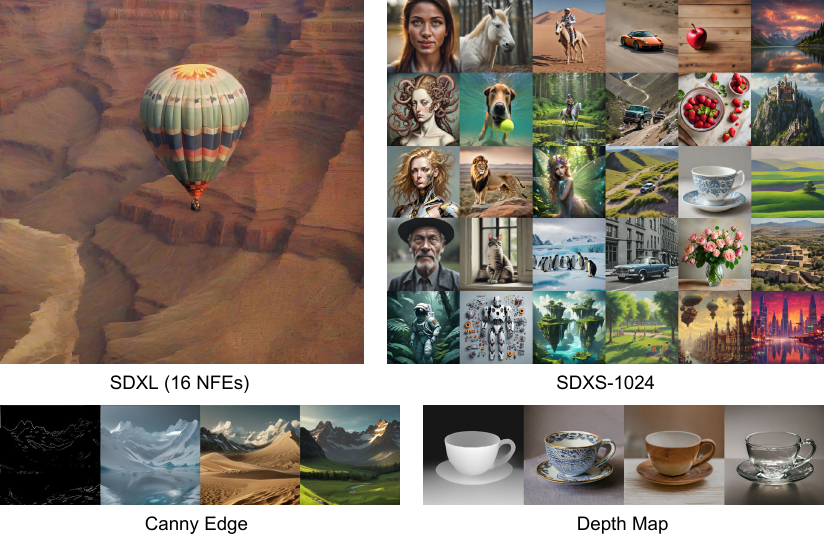}
    \caption{Assuming the image generation time is limited to \textbf{1 second}, then SDXL can only use 16 NFEs to produce a slightly blurry image, while SDXS-1024 can generate 30 clear images. Besides, our proposed method can also train ControlNet.}
    \label{fig:intro}
\end{figure*}

\begin{abstract}
    Recent advancements in diffusion models have positioned them at the forefront of image generation. Despite their superior performance, diffusion models are not without drawbacks; they are characterized by complex architectures and substantial computational demands, resulting in significant latency due to their iterative sampling process. To mitigate these limitations, we introduce a dual approach involving model miniaturization and a reduction in sampling steps, aimed at significantly decreasing model latency. Our methodology leverages knowledge distillation to streamline the U-Net and image decoder architectures, and introduces an innovative one-step DM training technique that utilizes feature matching and score distillation. We present two models, SDXS-512 and SDXS-1024, achieving inference speeds of approximately \textbf{100 FPS} ($30 \times$ faster than SD v1.5) and \textbf{30 FPS} ($60 \times$ faster than SDXL) on a single GPU, respectively. Moreover, our training approach offers promising applications in image-conditioned control, facilitating efficient image-to-image translation.
\end{abstract}

\section{Introduction}

Generative models have recently garnered substantial interest. Leveraging large-scale models and datasets, the text-to-image (T2I) models based on Diffusion Models (DM)~\cite{ldm,podell2023sdxl,dalle2,imagen} have exhibited remarkable capabilities in synthesizing images that are not only realistic but also closely adhere to textual prompts. Based on pretrained diffusion-based text-to-image models, a variety of extended applications have been developed, encompassing areas such as image editing~\cite{zhang2023adding,meng2021sdedit}, image inpainting~\cite{lugmayr2022repaint,yu2023inpaint}, image super-resolution~\cite{wang2023exploiting,yue2024resshift}, video generation~\cite{he2022latent,blattmann2023stable}, and 3D assets synthesis~\cite{poole2022dreamfusion,wang2024prolificdreamer}. Despite the rapid and diverse advancements, the deployment of DMs onto low-power devices, such as smartphones, remains a formidable challenge due to the models' large scale and the intrinsic nature of their multi-step sampling processes. Furthermore, even on cloud-based computing platforms equipped with high-performance GPUs, the considerable energy consumption are an aspect that warrants serious consideration. 

For common text-to-image DMs, the costs can be approximately calculated as the total latency, which includes the latencies of the text encoder, image decoder, and the denoising model, multiplied by the Number of Function Evaluations (NFEs). Efforts to mitigate the memory, computational, and storage overheads associated with DMs are underway. Similar to other large-scale model deployments, DMs can benefit from pruning~\cite{fang2024structural}, distillation~\cite{kim2023bk,gupta2024progressive}, and quantization~\cite{shang2023post,li2023q,he2024ptqd} to decrease their costs. Additionally, given the extensive use of Transformer layers within DMs, specific optimizations for Transformer layers~\cite{dao2022flashattention,xFormers2022} are capable of offering considerable efficiency improvements. Importantly, a significant research focus has been placed on minimizing the NFEs, given its profound influence on the overall latency of the model. Techniques employing progressive distillation~\cite{progressive,ondistillation,li2024snapfusion} and consistency distillation~\cite{song2023consistency,luo2023latent,luo2023lcm} have demonstrated the ability to lower NFEs to a range of 4 to 8. Furthermore, innovative strategies Rectified Flow-based~\cite{liu2023flow,liu2024instaflow} and Generative Adversarial Network (GAN)-based~\cite{xu2023ufogen,sauer2023adversarial,lin2024sdxl} methods have achieved reductions of NFEs to as low as 1. These advancements illuminate the promising potential for deploying diffusion models on edge devices. However, the simultaneous exploration of model miniaturization and the transition to one-step operations remains scarcely addressed within the literature. More importantly, although some methods can finetune the model into a few-steps model through LoRA~\cite{hu2021lora} and then directly use the original ControlNet~\cite{zhang2023adding} model for image-conditioned generation, this is suboptimal because the distribution of intermediate feature maps in the model updated by LoRA will still differ from that of the original model. Moreover, when the model needs to be finetuned to a one-step model, the low-rank update becomes insufficient for ensuring desired outcomes, and full finetuning would lead to even greater differences in feature map distribution. Therefore, we urgently need a method that can train ControlNet on a one-step model. 

\begin{table}[ht]
    \begin{minipage}{0.49\linewidth}
    \centering
    \captionof{table}{\small
    \textbf{Latency Comparison} between SD v2.1 base and our proposed efficient diffusion models on generating $512 \times 512$ images with batch size = 1.} \label{table:latency_analysis_1}
    \resizebox{1\linewidth}{!}{
    \begin{tabular}{r|ccccc}
    \toprule
    SD v2.1 base & Text Encoder & U-Net & Image Decoder \\
    \hline 
    \#Parameters & 0.33B & 0.87 B & 50 M \\ 
    Latency (ms) & 1.1 & 16 & 18 \\ 
    NFEs & 2 & 32 & 1 \\ 
    Total (ms) & 2.2 & 512 & 18 \\ \hline\hline 
    \textbf{SDXS-512} & Text Encoder & \textbf{U-Net} & \textbf{Image Decoder} \\
    \hline  
    \#Parameters & 0.33B & \textbf{0.32 B} & \textbf{1.2 M} \\ 
    Latency (ms) & 1.1 & \textbf{6.3} & \textbf{1.7} \\ 
    NFEs & \textbf{1} & \textbf{1} & 1 \\ 
    Total (ms) & 1.1 & \textbf{6.3} & \textbf{1.7} \\
    \bottomrule
    \end{tabular}
    }
    \end{minipage}\hfill
    \begin{minipage}{0.49\linewidth}
      \centering
      \captionof{table}{\small
      \textbf{Latency Comparison} between SDXL and our proposed efficient diffusion models on generating $1024 \times 1024$ images with batch size = 1.} \label{table:latency_analysis_2}
      \resizebox{1\linewidth}{!}{
      \begin{tabular}{r|ccccc}
      \toprule
      SDXL & Text Encoder & U-Net & Image Decoder \\
      \hline 
      \#Parameters & 0.80B & 2.56 B & 50 M \\ 
      Latency (ms) & 1.8 & 56 & 73 \\ 
      NFEs & 2 & 32 & 1 \\ 
      Total (ms) & 3.6 & 1792 & 73 \\ \hline\hline 
      \textbf{SDXS-1024} & Text Encoder & \textbf{U-Net} & \textbf{Image Decoder} \\
      \hline 
      \#Parameters & 0.80B & \textbf{0.74 B} & \textbf{1.2 M} \\ 
      Latency (ms) & 1.8 & \textbf{24} & \textbf{6.1} \\ 
      NFEs & \textbf{1}  & \textbf{1} & 1 \\ 
      Total (ms) & 1.8 & \textbf{24} & \textbf{6.1} \\
      \bottomrule
      \end{tabular}
      }
    \end{minipage}
\end{table}

In this paper, we provide a more comprehensive exploration of the aforementioned challenges. Initially, we focus on reducing the size of the VAE~\cite{kingma2013auto} decoder and U-Net~\cite{ronneberger2015unet}, both are resource-intensive components in DM sampling. We train an extremely lightweight image decoder to mimic the original VAE decoder’s output through a combination of output distillation loss and GAN loss. Following this, we leverage the block removal distillation strategy~\cite{kim2023bk} to efficiently transfer the knowledge from the original U-Net to a more compact version, effectively removing the majority of modules that contribute to latency. To reduce the NFEs, we propose a fast and stable training method. First, we suggest straightening the sampling trajectory and quickly finetuning the multi-step model into a one-step model by replacing the distillation loss function with the proposed feature matching loss. Then, we extend the Diff-Instruct training strategy~\cite{luo2024diff}, using the gradient of the proposed feature matching loss to replace the gradient provided by score distillation in the latter half of the timestep. We name our proposed method SDXS to admire SDXL~\cite{podell2023sdxl}, and provide two versions: $512\times512$ and $1024\times1024$, developed based on SD v2.1 base and SDXL, respectively. As shown in Tables~\ref{table:latency_analysis_1} and \ref{table:latency_analysis_2}, SDXS demonstrates efficiency far surpassing that of the base models, even achieving image generation at 100 FPS for $512\times512$ images and 30 FPS for $1024\times1024$ images on the GPU. Finally, to ease the application of our optimized model to tasks involving image-conditioned generation, we have adapted the block removal distillation strategy for ControlNet~\cite{zhang2023adding}.
Then, we extend our proposed training strategy to the training of ControlNet, relying on adding the pretrained ControlNet to the score function.

\section{Preliminary}

\subsection{Diffusion Models}

The forward process of DMs~\cite{ho2020denoising} transforms samples from the real distribution $p_0(\bm{x})$ into ones that follow a standard Gaussian distribution $\mathcal{N}(\mathbf{0}, \mathbf{I})$ by progressively adding noise. To counter this, the reverse process aims to invert the forward process by training a denoising model:
\begin{equation}\label{eqn:dm_loss}
    \mathcal{L}_{DM} = \mathbb{E}_{t \in [0, T],\bm{x}_0\sim p_0(\bm{x}),\bm{\epsilon}\sim\mathcal{N}(\mathbf{0}, \mathbf{I})} \; \lvert\lvert\hat{ \bm{\epsilon}}_{\bm{\theta}}(t, \mathbf{x}_t) -\bm{\epsilon} \rvert\rvert_2^2.
\end{equation}
This framework facilitates the generation of samples $\bm{\epsilon} \sim \mathcal{N}(\mathbf{0}, \mathbf{I})$, which, through the trained model, are transformed into new samples $\mathbf{x} \sim p_{\text{data}}(\mathbf{x})$ using step-by-step denoising.

Although DMs are derived from the Bayesian framework, they can also be considered as a special Variance Preserving (VP) case under the Score SDE framework~\cite{song2020score}, which simulates the continuous dynamic changes in the data generation process. At its core, the loss function employed is the score matching (SM) loss, aimed at minimizing the difference between the model's estimated score and the true score of the data:
\begin{equation}\label{eqn:SM}
    \mathcal{L}_{SM} = \int_{t=0}^T w(t) \mathbb{E}_{\bm{x}_0\sim p_0(\bm{x}), \bm{x}_t|\bm{x}_0 \sim p_t(\bm{x}_t|\bm{x}_0)} \|\bm{s}_\phi(\bm{x}_t,t) - \nabla_{\bm{x}_t}\log p_t(\bm{x}_t|\bm{x}_0)\|_2^2\mathrm{d}t.
\end{equation}
In practice, while we still utilize DM framework and its discrete training process, the underlying principle for the loss function remains consistent with the continuous interpretation offered by the Score SDE framework.

\subsection{Diff-Instruct}
Although DMs have demonstrated their capability to produce high-quality images, their efficiency is hindered by the requirement for multiple steps during the sampling process. It is necessary to seek a method to distill its accumulated knowledge from pretrained DMs to instruct the training of models capable of generating samples in just one step. This knowledge distillation from pretrained DMs is known as score distillation and was first proposed in the field of 3D assets synthesis~\cite{poole2022dreamfusion}. Diff-Instruct~\cite{luo2024diff} brings score distillation back into image generation, relying on the definition of Integral Kullback-Leibler (IKL) divergence between two distributions $p,q$:
\begin{equation}\label{eqn:ikl1}
    \mathcal{D}_{IKL}^{[0,T]}(q,p) = \int_{t=0}^T w(t)\mathbb{E}_{\bm{x}_t\sim q_{t}(\bm{x})}\big[ \log \frac{q_{t}(\bm{x}_t)}{p_{t}(\bm{x}_t)}\big]\mathrm{d}t,
\end{equation}
where $q_t$ and $p_t$ denote the marginal densities of the diffusion process at time $t$.
The gradient of the IKL in (\ref{eqn:ikl1}) between $q_0$ and $p_0$ is
\begin{equation}\label{eqn:ikl_grad11}
\operatorname{Grad}(\theta)=\int_{t=0}^T w(t)\mathbb{E}_{\bm{x}_0 = g_\theta(\bm{z}), \bm{x}_t|\bm{x}_0 \sim p_t(\bm{x}_t|\bm{x}_0)}\big[ \bm{s}_{\phi}(\bm{x}_t,t) - \bm{s}_{p_t}(\bm{x}_t)\big]\frac{\partial \bm{x}_t}{\partial\theta}\mathrm{d}t.
\end{equation}
where $\bm{x}_0 = g_\theta(\bm{z})$ denotes the sample generated by the one-step generator being trained to accept a randomly initialized latent $z$, and $\bm{s}_{\phi}$ and $\bm{s}_{p_t}$ denote the score functions of the DM trained online on the generated data and the pretrained DM, respectively. Diff-Instruct uses this gradient directly to update the generator, and when the outputs of the two score functions agree, the marginal distribution of the one-step generator output is consistent with the marginal distribution of the pretrained DM. It’s worth noting that the core concept of Diff-Instruct closely aligns with that of VSD~\cite{wang2024prolificdreamer}, and this concept has also been recently expanded to text-to-image generation tasks~\cite{yin2023one,nguyen2023swiftbrush}.

\section{Method}

\subsection{Architecture Optimizations}

The image generation process in the Latent Diffusion Model (LDM) framework consists of three key elements: a text encoder, an image decoder, and a denoising model that requires multiple iterations for a clear image. Given the relatively low overhead associated with the text encoder, optimizing its size has not been deemed a priority. 

\begin{figure*}[t]
    \centering
    \includegraphics[width=\linewidth]{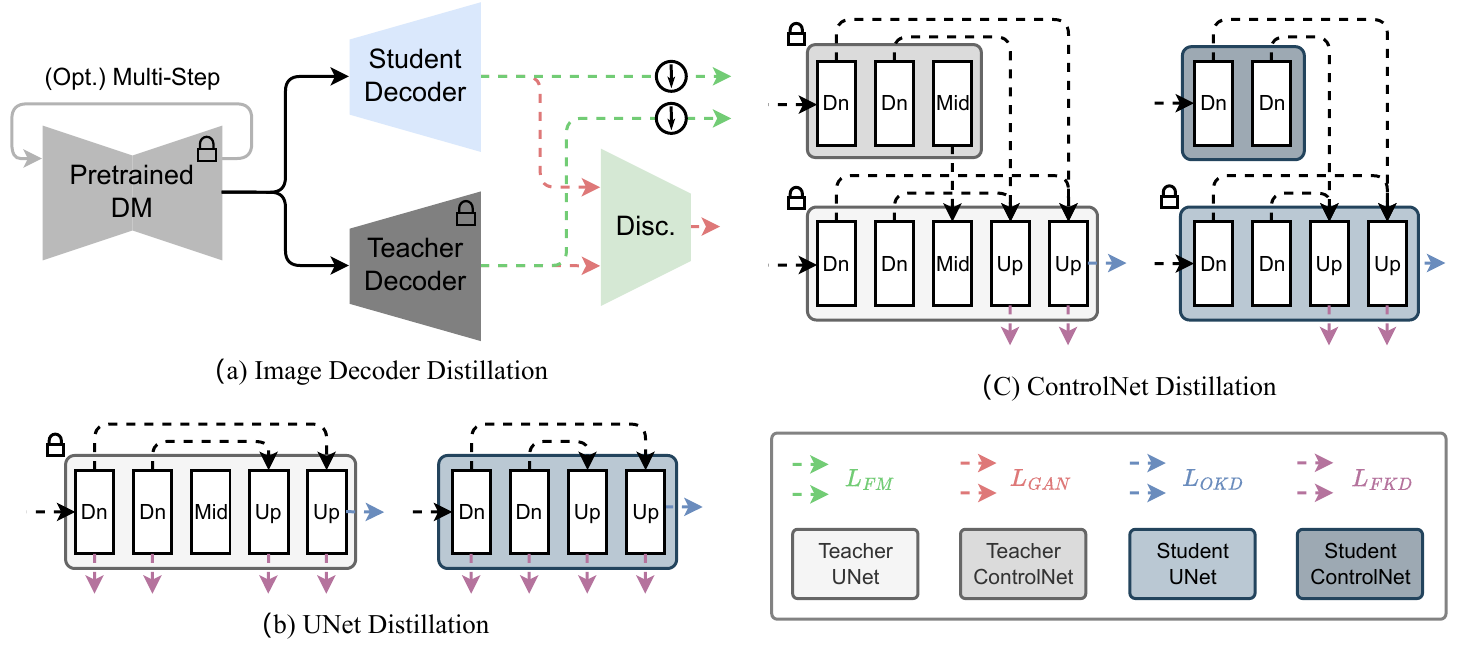}
    \caption{Network architecture distillation, including image decoder, U-Net and ControlNet.}
    \label{fig:arch}
\end{figure*}

\paragraph{VAE.}
The LDM framework significantly improves the training efficiency for high-resolution image diffusion models by projecting samples to a more computationally efficient lower-dimensional latent space. This is facilitated through high-ratio image compression using pretrained models, such as the Variational AutoEncoder (VAE)~\cite{kingma2013auto,rezende2014stochastic} or the Vector Quantised-Variational AutoEncoder (VQ-VAE)~\cite{van2017neural,esser2021taming}. The VAE, in particular, includes an encoder to map images into latent space and a decoder to reconstruct images. Its training is optimized by balancing three losses: reconstruction, Kullback-Leibler (KL) divergence, and GAN loss. However, equally treating all samples during training introduces redundancy. Utilizing a pretrained diffusion model $F$ to sample latent codes \(\bm{z}\) and a pretrained VAE decoder to reconstruct images \(\tilde{\bm{x}}\), we introduce a VAE Distillation (VD) loss for training a tiny image decoder \(G\):
\begin{equation}\label{eqn:VAED}
    \mathcal{L}_{VD} = \| G(\bm{z})_{\downarrow 8\times} - \tilde{\bm{x}}_{\downarrow 8\times}\|_1 + \lambda_{GAN} {\mathcal{L}_{GAN}(G(\bm{z}), \tilde{\bm{x}}, D)},
\end{equation}
where \(D\) is the GAN discriminator, $\lambda_{GAN}$ is used to balance two loss terms, and $\| G(\bm{z})_{\downarrow \times 8} - \bm{x}_{\downarrow \times 8}\|_1$ means the $L_1$ loss is measured on $8\times$ downsampled images . Figure~\ref{fig:arch} (a) illustrates the training strategy for distilling tiny image decoder. We further advocate for a streamlined CNN architecture devoid of complex components like attention mechanisms and normalization layers, focusing solely on essential residual blocks and upsampling layers.

\paragraph{U-Net.}
LDMs employ U-Net architectures~\cite{ronneberger2015unet}, incorporating both residual and Transformer blocks, as their core denoising model. To leverage the pretrained U-Nets' capabilities while reducing computational demands and parameter numbers, we adopt a knowledge distillation strategy inspired by the block removal training strategy from BK-SDM~\cite{kim2023bk}. This involves selectively removing residual and Transformer blocks from the U-Net, aiming to train a more compact model that can still reproduce the original model's intermediate feature maps and outputs effectively. Figure~\ref{fig:arch} (b) illustrates the training strategy for distilling tiny U-Net. The knowledge distillation is achieved through output knowledge distillation (OKD) and feature knowledge distillation (FKD) losses:
\begin{equation}
    \mathcal{L}_{OKD} = \int_{t=0}^T \mathbb{E}_{\bm{x}_0\sim p_0(\bm{x}), \bm{x}_t|\bm{x}_0 \sim p_t(\bm{x}_t|\bm{x}_0)} \|\bm{s}_\theta(\bm{x}_t,t) - \bm{s}_\phi(\bm{x}_t,t)\|_2^2\mathrm{d}t,
\end{equation}
\begin{equation}
    \mathcal{L}_{FKD} = \int_{t=0}^T \mathbb{E}_{\bm{x}_0\sim p_0(\bm{x}), \bm{x}_t|\bm{x}_0 \sim p_t(\bm{x}_t|\bm{x}_0)} \sum_{l}  \|\bm{f}_\theta^l(\bm{x}_t,t) - \bm{f}_\phi^l(\bm{x}_t,t)\|_2^2\mathrm{d}t,
\end{equation}
with the overarching loss function being a combination of the two:
\begin{equation}
    \mathcal{L}_{KD} = \mathcal{L}_{OKD} + \lambda_{F} \mathcal{L}_{FKD},
\end{equation}
where \(\lambda_{F}\) balances the two loss terms. Different from BK-SDM, we exclude the original denoising loss. Our models are miniaturized based on SD-2.1 base and SDXL-1.0 base. For SD-2.1 base, we remove the middle stage, the last stage of the downsampling stages and the first stage of the upsampling stages, and remove the Transformer blocks of the highest resolution stages. For SDXL-1.0 base, we remove most of the Transformer blocks.

\paragraph{ControlNet.}
ControlNet~\cite{zhang2023adding} boosts diffusion models by embedding spatial guidance in existing text-to-image frameworks, enabling image-to-image tasks like sketch-to-image translation, inpainting, and super-resolution. It copys U-Net's encoder architecture and parameters, adding extra convolutional layers to incorporate spatial controls. Despite inheriting U-Net's parameters and employing zero convolutions for enhanced training stability, ControlNet's training process remains expensive and is significantly affected by the dataset quality. To address these challenges, we propose a distillation approach that distills the ControlNet of the original U-Net into the corresponding ControlNet of the tiny U-Net. As illustrated in Figure~\ref{fig:arch} (b), instead of directly distilling the output of ControlNet's zero convolutions, we combine ControlNet with U-Net and then distill the intermediate feature maps and output of U-Net, which allows the distilled ControlNet and the tiny U-Net to work better together. Considering that ControlNet does not affect the feature map of U-Net's encoder, feature distillation is only applied to U-Net's decoder. 

\subsection{One-Step Training}

\begin{figure*}[t]
    \centering
    \includegraphics[width=\linewidth]{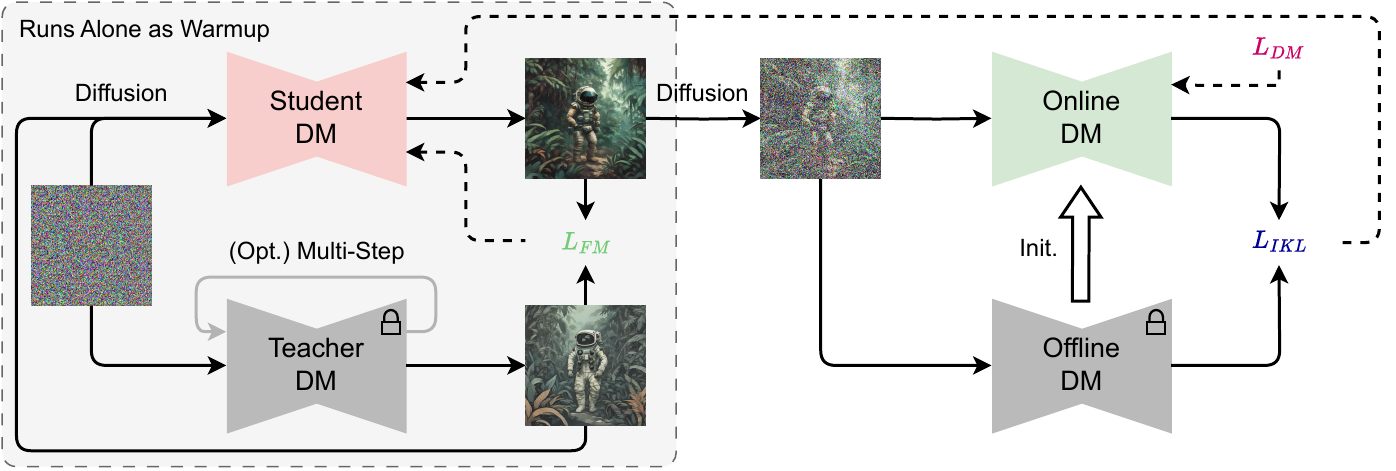}
    \caption{The proposed one-step U-Net training strategy based on feature matching and score distillation. The dashed lines indicate the gradient backpropagation.}
    \label{fig:main}
\end{figure*}

While DMs excel in image generation, their reliance on multiple sampling steps introduces significant inference latency even with advanced samplers~\cite{ddim,liu2022pseudo,lu2022dpm,lu2022dpmplus,zhao2024unipc}. To address this, prior studies have introduced knowledge distillation techniques, such as progressive distillation~\cite{progressive,ondistillation,li2024snapfusion} and consistency distillation~\cite{song2023consistency,luo2023latent,luo2023lcm}, aiming to reduce the sampling steps and accelerate inference. However, these approaches typically can only produce clear images with 4$\sim$8 sampling steps, which starkly contrasts the one-step generation process seen in GANs. Exploring the integration of GANs into the DM training regime has shown promise for enhancing image quality~\cite{xu2023ufogen,sauer2023adversarial}. However, GANs come with their own set of challenges, including sensitivity to hyperparameters and training instabilities. It is necessary to seek a more stable training strategy for one-step generation models.

\begin{figure*}[ht]
  \centering
  \includegraphics[width=\linewidth]{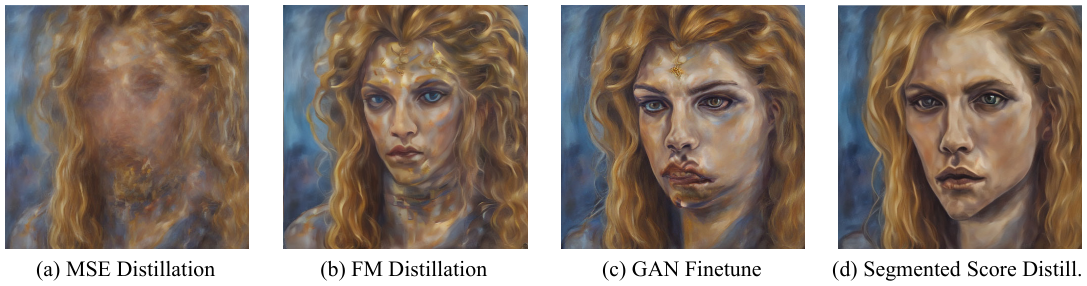}
  \caption{Comparison of images generated by models trained with different distillation strategies. Prompt: Self-portrait oil painting, a beautiful woman with golden hair.}
  \label{fig:ablation}
\end{figure*}

\paragraph{Feature Matching Warmup.}
A straightforward approach to training a one-step model involves initializing noises $\bm{\epsilon}$ and employing an Ordinary Differential Equation (ODE) sampler $\psi$ to sample and obtain the generated images $\bm{\hat{x}}_0$, thereby constructing noise-image pairs. These pairs then serve as inputs and ground truth for the student model during training. This method, however, often results in the production of low-quality images. The underlying issue is the crossings in the sampling trajectories of noise-image pairs generated using the ODE sampler from a pretrained DM, leading to an ill-posed problem. Rectified Flow~\cite{liu2023flow,liu2024instaflow} tackles this challenge by straightening the sampling trajectories. It replaces the training objective and proposes a 'reflow' strategy to refine the pairing, thus minimizing trajectory crossings. Instead, we note that the crossing of sampling trajectories can lead to a one noise input corresponding to multiple ground-truth images, causing the trained model generating an image that is a weighted sum of multiple feasible outputs with weights $w(y)$:
\begin{equation}\label{eqn:reweight}
    \min_{\hat{y}} \mathbb{E}[\mathcal{L}(y, \hat{y})] \Rightarrow \hat{y} = \int w(y) \cdot y \, dy .
\end{equation}
For the most commonly used mean square error (MSE) loss, while consisting of multiple feasible targets, the model tends to output the average of multiple feasible solutions to minimize the overall error, which in turn leads to the blurriness of the generated images. To address this, we explore alternative loss functions that alter the weighting scheme to prioritize sharper images. At the most cases, we can use L1 loss, perceptual loss~\cite{johnson2016perceptual}, and LPIPS loss~\cite{zhang2018unreasonable} to change the form of weighting. We build upon the method of feature matching~\cite{salimans2016improved}, which involves computing the loss on the intermediate feature maps generated by a encoder model. Specifically, we draw inspiration from the DISTS loss~\cite{ding2020image} to apply the Structural Similarity Index (SSIM) on these feature maps for a more refined feature matching loss:
\begin{equation}
    \mathcal{L}_{FM} = \sum_{l} w_l \cdot \text{SSIM}(\bm{f}_\theta^l(\mathbf{x}_\theta(\bm{\epsilon})), \bm{f}_\theta^l(\psi (\mathbf{x}_\phi(\bm{\epsilon})))).
\end{equation}
where $w_l$ is the weight of the SSIM loss calculated on the $l$-th intermediate feature map encoded by the encoder $\bm{f}_\theta$, $\mathbf{x}_\theta(\bm{\epsilon})$ is the image generated by tiny U-Net $\mathbf{x}_\theta$, and $\psi (\mathbf{x}_\phi(\bm{\epsilon}))$ is the image generated by original U-Net $\mathbf{x}_\phi$ with ODE sampler $\psi$. In practice, we find that using the pretrained CNN backbone~\cite{simonyan2014very}, ViT backbone~\cite{oquab2023dinov2}, and the encoder of DM U-Net all yield favorable results, with a comparison to MSE loss shown in Figure~\ref{fig:ablation}. Besides, we also straighten the model's trajectories to narrow the range of feasible outputs using existing finetuning methods like LCM~\cite{luo2023latent,luo2023lcm} or directly use the publicly available few-step models~\cite{sauer2023adversarial,lin2024sdxl}. We will use $\mathcal{L}_{FM}$ alone to train the one-step model as a warmup, relying only on a small number of training steps.

\begin{algorithm}[t]
    \SetAlgoLined
    \KwIn{offline DM $\bm{s}_{p_t}$, online DM $\bm{s}_\phi$, ODE sampler $\psi$, prior distribution $p_z$, one-step DM $\mathbf{x}_\theta$, hyperparameters $\lambda_{FM}$ and $\alpha$.}
    \While{not converge}{
    ~~update ${\phi}$ using SGD with gradient
    $$\operatorname{Grad}(\phi) = \frac{\partial}{\partial\phi}\int_{t=0}^T w(t) \mathbb{E}_{\mathbf{z}\sim p_z, \bm{x}_0 = \mathbf{x}_\theta(\mathbf{z}),\atop \bm{x}_t|\bm{x}_0 \sim p_t(\bm{x}_t|\bm{x}_0)} \|\bm{s}_\phi(\bm{x}_t,t) - \nabla_{\bm{x}_t}\log p_t(\bm{x}_t|\bm{x}_0)\|_2^2\mathrm{d}t.$$
    update $\theta$ using SGD with the gradient
    \begin{align*}
        \operatorname{Grad}(\theta) & = \int_{t=0}^{\alpha T} w(t)\mathbb{E}_{\bm{x}_0 = \mathbf{x}_\theta(\bm{z}), \bm{x}_t|\bm{x}_0 \sim p_t(\bm{x}_t|\bm{x}_0)}\big[ \bm{s}_{\phi}(\bm{x}_t,t) - \bm{s}_{p_t}(\bm{x}_t)\big]\frac{\partial \bm{x}_t}{\partial\theta} \mathrm{d}t \\
        & + \lambda_{FM} \frac{\partial \mathcal{L}_{FM}(\mathbf{x}_\theta(\bm{x}_t), \psi (\mathbf{x}_\phi(\bm{x}_t)))}{\partial\theta}.
    \end{align*}
    adjust hyperparameters $\lambda_{FM}$ and $\alpha$.
    }
    \Return{$\theta,\phi$.}
    \caption{Segmented Score Distillation Algorithm}
    \label{alg:algo1}
\end{algorithm}

\paragraph{Segmented Score Distillation.}
Although feature matching loss can produce almost clear images, it falls short of achieving an true distribution match, so the trained model can only be used as an initialization for formal training. To address this gap, we elaborate on the training strategy utilized in Diff-Instruct, which aims to align the model's output distribution more closely with that of a pretrained model by matching marginal score functions over timestep. However, because it requires adding high levels of noise at $t \rightarrow T$ for the target score to be calculable, the score function estimated at this time is inaccurate~\cite{song2019generative,alldieck2024score}. We note that the sampling trajectory of diffusion models from coarse to fine, which means that $t \rightarrow T$, the score function provides gradients of low-frequency information, while $t \rightarrow 0$, it offers gradients of high-frequency information. Therefore, we divide the timestep into two segments: $[0, \alpha T]$ and $(\alpha T, T]$, with the latter being replaced by $\mathcal{L}_{FM}$ because it can provide sufficient low-frequency gradients. This strategy can be formally represented as:
\begin{equation}\label{eqn:sikl_grad}
    \begin{split}
        \operatorname{Grad}(\theta) & = \frac{\partial \mathcal{L}_{IKL}(t, \bm{x}_t, \bm{s}_{\phi})}{\partial\theta} = \int_{t=0}^{\alpha T} F(t, \bm{x}_t)\mathrm{d}t + \int_{t=\alpha T}^T F(t, \bm{x}_t)\mathrm{d}t \\
        & \approx \int_{t=0}^{\alpha T} F(t, \bm{x}_t)\mathrm{d}t + \lambda_{FM} \frac{\partial \mathcal{L}_{FM}(\mathbf{x}_\theta(\bm{x}_t), \psi (\mathbf{x}_\phi(\bm{x}_t)))}{\partial\theta},
    \end{split}
\end{equation}
where
\begin{equation}
    \begin{split}
        F(t, \bm{x}_t) = w(t)\mathbb{E}_{\bm{x}_0 = \mathbf{x}_\theta(\bm{z}), \bm{x}_t|\bm{x}_0 \sim p_t(\bm{x}_t|\bm{x}_0)}\big[ \bm{s}_{\phi}(\bm{x}_t,t) - \bm{s}_{p_t}(\bm{x}_t)\big]\frac{\partial \bm{x}_t}{\partial\theta},
    \end{split}
\end{equation}
$\lambda_{FM}$ is used to balance gradients of the two segments, and \(\alpha \in [0, 1]\). we intentionally set \(\alpha\) close to 1 and \(\lambda_{FM}\) at a high value to ensure the model's output distribution smoothly aligns the predicted distribution by the pretrained score function. After achieving significant overlap in the probability densities, we gradually lower both \(\alpha\) and \(\lambda_{FM}\). Figure~\ref{fig:main} visually depicts our training strategy, where the offline DM represents the U-Net of a pretrained DM, and the online DM is initialized from the offline DM and finetuned on the generated images through Eq.~\eqref{eqn:dm_loss}. In practice, the online DM and student DM are trained alternately as in Algorithm \ref{alg:algo1}.

\begin{figure*}[t]
    \centering
    \includegraphics[width=\linewidth]{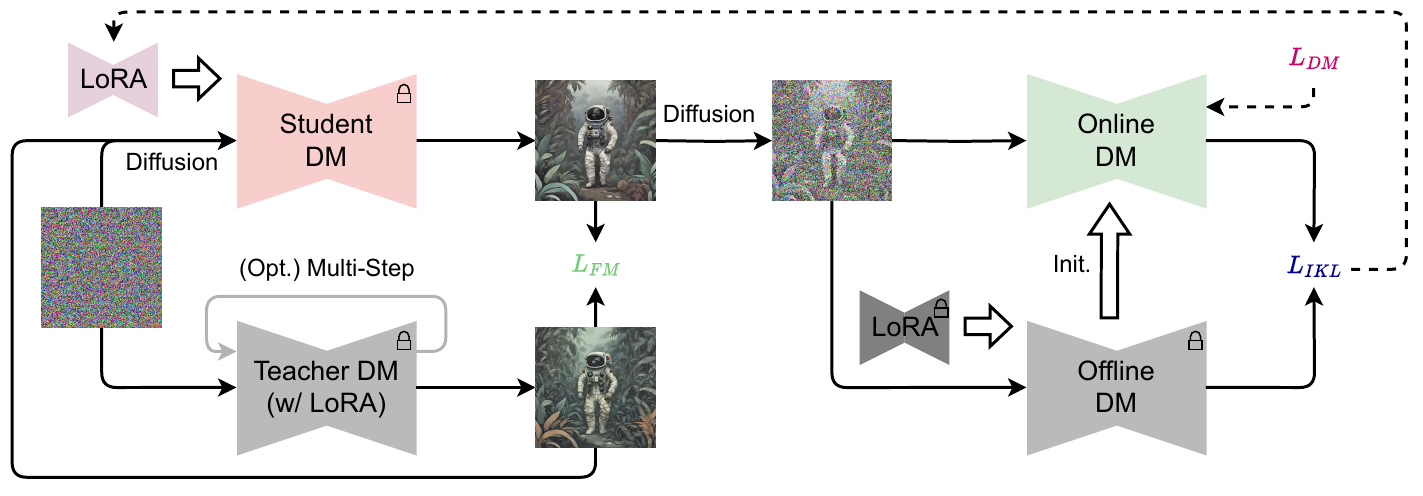}
    \caption{The proposed LoRA training strategy based on feature matching and score distillation. The dashed lines indicate the gradient backpropagation.}
    \label{fig:lora}
\end{figure*}

\paragraph{LoRA.} 
Once the one-step DM is trained, it can be finetuned like other DMs to adjust the style of the generated images. We utilize LoRA~\cite{hu2021lora} in conjunction with the proposed Segmented Score Distillation to finetune an one-step DM, as illustrated in Figure~\ref{fig:lora}. Specifically, we insert the pretrained LoRA into the offline DM, and if it is also compatible with the Teacher DM, it is inserted there as well. It is important to note that we do not insert LoRA into the online DM, as it corresponds to the output distribution of the one-step DM. Then, we use the same training procedure as we do for one-step training, but skip the feature matching warmup, since LoRA finetune is much more stable than full finetune. Besides, when the Teacher DM cannot incorporate the pretrained LoRA, we use a reduced \(\lambda_{FM}\). In this way, the pretrained LoRA can be distilled into the SDXS’s LoRA.

\paragraph{ControlNet.}

\begin{figure*}[t]
    \centering
    \includegraphics[width=\linewidth]{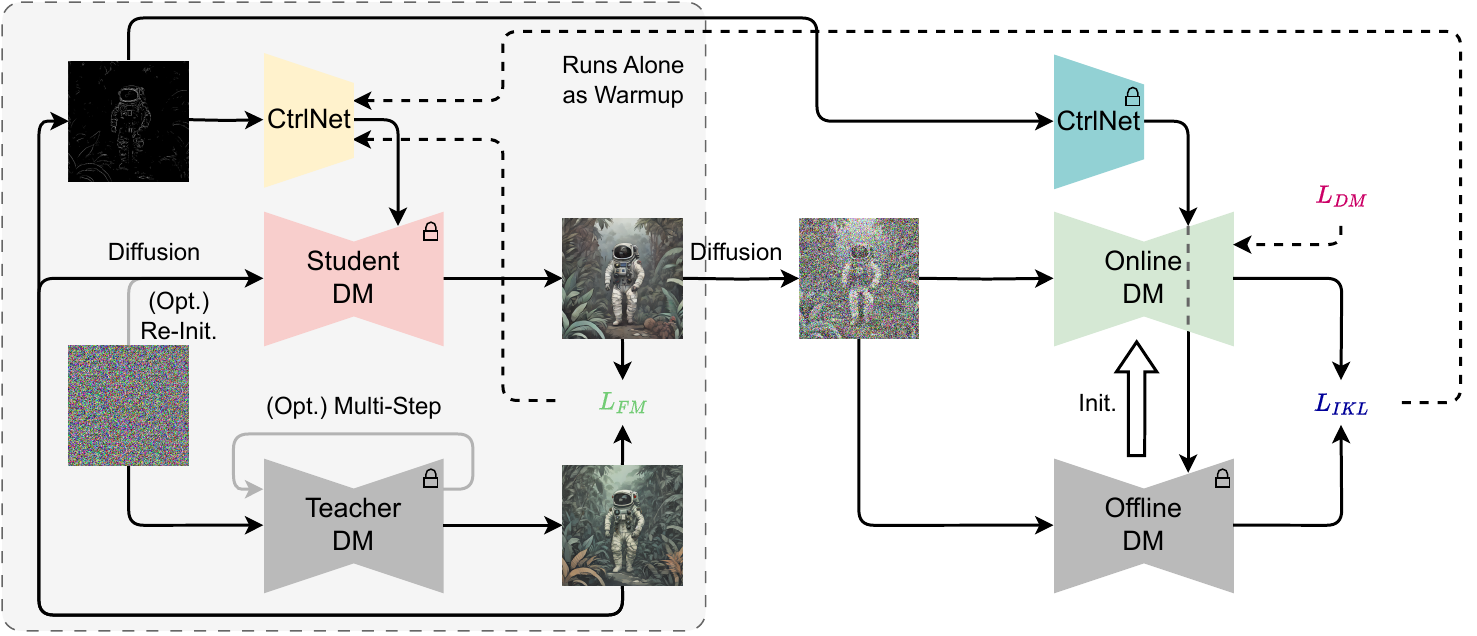}
    \caption{The proposed one-step ControlNet training strategy based on feature matching and score distillation. The dashed lines indicate the gradient backpropagation.}
    \label{fig:control}
\end{figure*}

Our approach can also be adapted for training ControlNet, enabling the tiny one-step model to incorporate image conditions into its image generation process, as depicted in Figure~\ref{fig:control}. Compared to the base model for text-to-image generation, the model trained here is the distilled ControlNet that accompanies the tiny U-Net mentioned earlier, and the parameters of the U-Net are fixed during training. Importantly, we need to extract the control images from the images sampled by the teacher model, rather than from the dataset images, to ensure that the noise, target image, and control image form a pairing triplet. Furthermore, the original multi-step U-Net's accompanying pretrained ControlNet is integrated with both the online U-Net and offline U-Net but does not participate in training. Similar to the text encoder, the function is confined to serving as a pretrained feature extractor. In this way, to further reduce \(\mathcal{L}\), the trained ControlNet is to learn to utilize the control images extracted from the target images. At the same time, the score distillation encourages the model to match the marginal distributions, enhancing the contextual relevance of the generated images. Notably, we found that replacing a portion of the noise input to U-Net with freshly reinitialized noise can enhance control capabilities.

\section{Experiment}

\textbf{Implementation Details.} Our code is developed based on diffusers library\footnote{\url{https://github.com/huggingface/diffusers}}. Because we cannot access the training datasets of SD v2.1 base and SDXL, the entire training process is almost data-free, relying solely on prompts available from publicly accessible dataset~\cite{schuhmann2022laion}. When necessary, we use open-source pretrained models in conjunction with these prompts to generate corresponding images. To train our model, we configure the training mini-batch size to range from 1,024 to 2,048. To accommodate this batch size on the available hardware, we strategically implement gradient accumulation when necessary. It's important to note that we found that the proposed training strategy results in models generating images with less texture. Therefore, after training, we utilize GAN loss in conjunction with extremely low-rank LoRA for a short period of fine-tuning. When GAN loss is needed, we use Projected GAN loss from StyleGAN-T~\cite{sauer2023stylegan}, and the basic settings are consistent with ADD~\cite{sauer2023adversarial}. For the training of SDXS-1024, we use Vega~\cite{gupta2024progressive}, a compact version of SDXL, as the initialization for both online DM and offline DM to reduce training overhead. 

\begin{table}[ht]
  \centering
  \captionof{table}{\small
  \textbf{Performance Comparison} on MS-COCO 2017 5K subset. If CFG~\cite{cfg} is enabled, then the scale will be set to 7.5, and NFEs will be twice the number of sampling steps. Latency is tested under the conditions of enabling float16, with a batch size of 1, and after compiling the model.} \label{table:performance_analysis}
  \resizebox{1\linewidth}{!}{
  \begin{tabular}{l|cccc|cccc} 
  \hline 
  Method & Resolution & \#Params of U-Net & Sampler & NFEs & Latency (ms) $\downarrow$ & FID $\downarrow$  & CLIP Score $\uparrow$  \\
  \hline 
  SD v1.5~\cite{ldm} & $512 \times 512$ & 860 M & DPM-Solver++  & 16 & 276 & 24.28 & 31.84 \\
  SD v1.5-LCM~\cite{luo2023lcm} & $512 \times 512$ & 860 M  & LCM & 4 & 84 & 34.74 & 30.85 \\
  SD Turbo~\cite{sauer2023adversarial} & $512 \times 512$ & 865 M  & - & 1 & 35 & 26.50 & 33.07 \\
  Tiny SD~\cite{kim2023bk} & $512 \times 512$ & 324 M  & DPM-Solver++ & 16 & 146 & 31.16 & 31.06 \\ 
  \rowcolor{gray!20} SDXS-512 (ours) & $512 \times 512$ & 319 M & - & 1 & 9 & 28.21 & 32.81 \\
  \hline 
  \hline 
  Method & Resolution & \#Params of U-Net & Sampler & NFEs & Latency (ms) $\downarrow$ & FID $\downarrow$  & CLIP Score $\uparrow$  \\
  \hline 
  SDXL~\cite{podell2023sdxl} & $1024 \times 1024$ & 2.56B & Euler & 32 & 1869 & 24.60 & 33.77 \\ 
  SDXL-LCM~\cite{luo2023lcm} & $1024 \times 1024$ & 2.56B & LCM & 4 & 299 & 28.87 & 32.45 \\
  SDXL Lightning~\cite{lin2024sdxl} & $1024 \times 1024$ & 2.56B & - & 1 & 131 & 29.26 & 32.09 \\
  Vega~\cite{gupta2024progressive} & $1024 \times 1024$ & 0.74B & Euler & 32 & 845 & 29.24 & 32.99 \\ 
  \rowcolor{gray!20} SDXS-1024 (ours) & $1024 \times 1024$  & 0.74B & - & 1 & 32 & 30.92 & 32.32 \\
  \hline 
  \end{tabular}
  }
\end{table}

\begin{figure*}[p]
  \centering
  \includegraphics[width=0.96\linewidth]{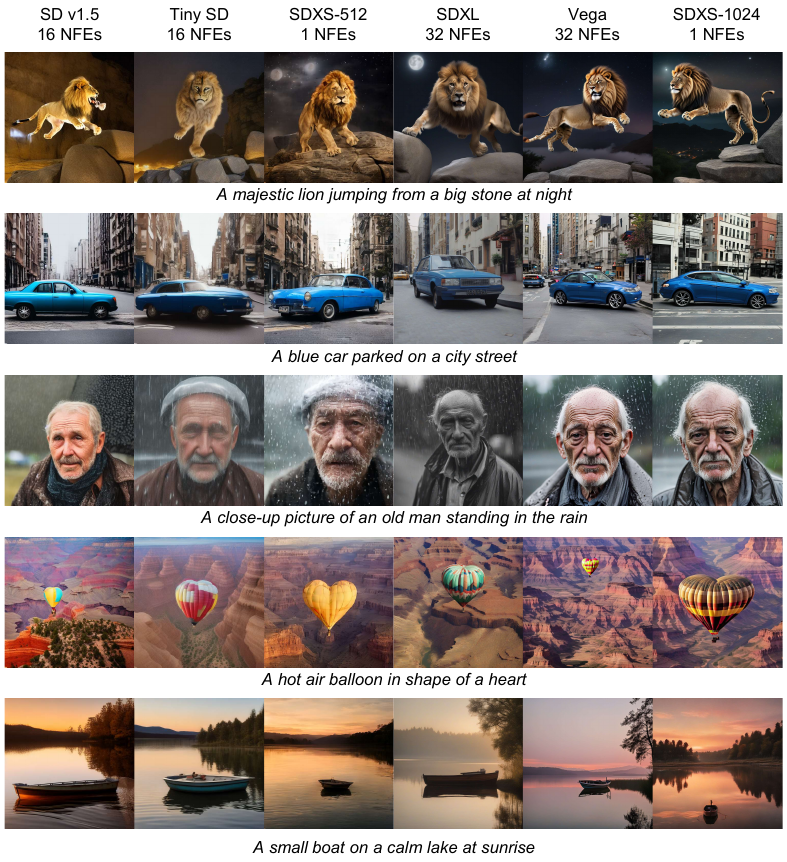}
  \caption{Qualitative comparison between SD v1.5, Tiny SD, SDXL, Vega, and our SDXS.}
  \label{fig:comp1}
\end{figure*}

\begin{figure*}[p]
  \centering
  \includegraphics[width=0.96\linewidth]{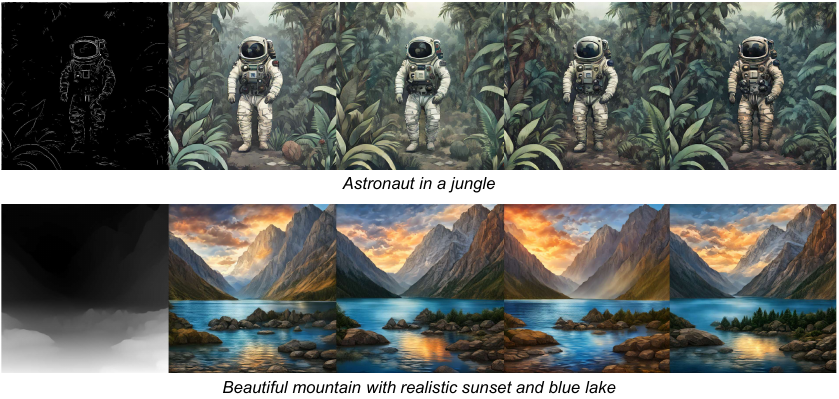}
  \caption{Two examples of ControlNet with SDXS-512.}
  \label{fig:comp2}
\end{figure*}

\subsection{Text-to-Image Generation}

\begin{figure*}[t]
  \centering
  \includegraphics[width=1.0\linewidth]{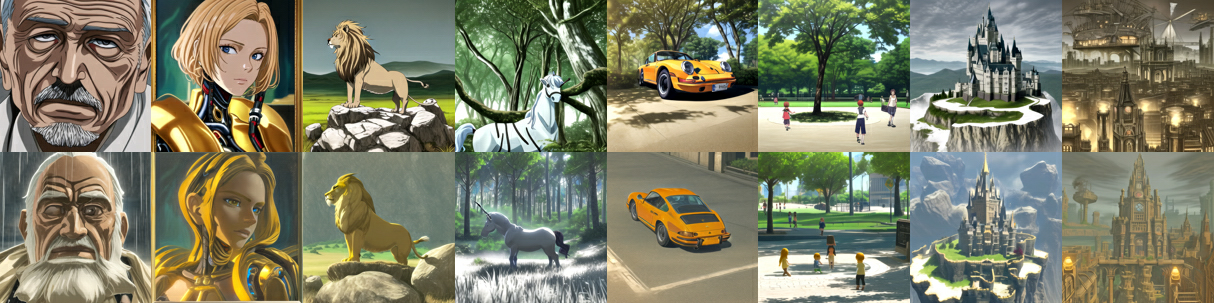}
  \caption{Two samples of two different styles of LoRA with top and bottom images generated using the same prompts.}
  \label{fig:comp_lora}
\end{figure*}

We report the quantitative results, \textit{i.e.}, FID~\cite{heusel2017gans} and CLIP scores~\cite{radford2021learning}, on MS-COCO 2017 validation set~\cite{mscoco} for evaluation. Due to the strong assumption of Gaussian distribution, FID is not a good indicator for measuring image quality~\cite{podell2023sdxl}, as it is significantly affected by the diversity of generated samples, but we still report it as prior works do. Table~\ref{table:performance_analysis} shows the performance comparison on MS-COCO 2017 5K subset and Figure~\ref{fig:comp1} shows some examples. Despite a noticeable downsizing in both the sizes of the models and the number of sampling steps required, the prompt-following capability of SDXS-512 remains superior to that of SD v1.5. Moreover, when compared to Tiny SD, another model designed for efficiency, the superiority of SDXS-512 becomes even more pronounced. This observation is consistently validated in the performance of SDXS-1024 as well.  Besides, we demonstrate the sample of training LoRA using the proposed method, as shown in Figure~\ref{fig:comp_lora}. It is evident that the style of the generated images by the model can be effectively transferred to match the style of the offline DM into which the style-oriented LoRA has been integrated, while generally maintaining the consistency of the scene's layout. 

\subsection{Image-to-Image Translation}

As we have illustrated earlier, our introduced one-step training approach is versatile enough to be applied to image-conditioned generation. Here, we demonstrate its efficacy in facilitating image-to-image conversions utilizing ControlNet, specifically for transformations involving canny edges and depth maps. Figure~\ref{fig:comp2} illustrates a representative example from each of two distinct tasks, highlighting the capability of the generated images to closely adhere to the guidance provided by control images. However, it also reveals a notable limitation in terms of image diversity. As shown in Figure~\ref{fig:intro}, while the problem can be mitigated by replacing the prompt, it still underscores an area ripe for enhancement in our subsequent research efforts.

\section{Conclusion}

This paper explores the distillation of large-scale diffusion-based text-to-image generation models into efficient versions that enable real-time inference on GPUs. Initially, we employ knowledge distillation to compress both the U-Net architecture and the image decoder. Subsequently, we introduce a novel training strategy that leverages feature matching and score distillation to reduce the sampling process to one step. This approach allows for the real-time generation of $1024 \times 1024$ images on a single GPU, maintaining quality comparable to original models. Moreover, the training methodology we propose can also adapt to tasks involving image-conditioned generation, eschewing the direct adaptation of the pretrained ControlNet. We believe that the deployment of efficient image-conditioned generation on edge devices represents a promising avenue for future research, with plans to explore additional applications such as inpainting and super-resolution.

{
\small
\bibliographystyle{unsrt}
\bibliography{neurips}
}
\end{document}